\documentclass[conference]{IEEEtran}
\usepackage{times}
\usepackage[numbers]{natbib}
\usepackage{multicol}
\usepackage[bookmarks=true]{hyperref}
\usepackage{graphics} 
\usepackage{epsfig} 
\usepackage{times} 
\usepackage{amsmath} 
\usepackage{amssymb}  
\usepackage{amsthm}
\usepackage{hyperref}
\usepackage{multirow}
\usepackage{caption,setspace}
\usepackage{booktabs}
\usepackage{arydshln}
\usepackage{tablefootnote}
\usepackage{csquotes}
\usepackage{footnote}
\makesavenoteenv{tabular}
\makesavenoteenv{table}
\usepackage[]{threeparttable}
\usepackage{breqn}
\usepackage{subcaption}
\usepackage{algorithm, algpseudocode}
\usepackage{float}
\usepackage[dvipsnames]{xcolor}
\let\labelindent\relax
\usepackage{enumitem}
\usepackage{tikz}

\usepackage{graphicx}  
\usepackage{subcaption}  

\DeclareMathOperator*{\argmin}{arg\,min}

\begin{document}

\title{Scalable Coverage Trajectory Synthesis on GPUs\\as Statistical Inference}

\author{%
  \IEEEauthorblockN{%
    Max M. Sun, \ 
    Jueun Kwon, \
    Todd Murphey %
  }%
  \IEEEauthorblockA{%
    Center for Robotics and Biosystems, Northwestern University, Evanston, IL 60208 \\
    Email: msun@u.northwestern.edu\\
    Project website: \textcolor{blue}{\url{https://murpheylab.github.io/lqr-flow-matching/}}
  }%
}

\maketitle
\allowdisplaybreaks

\begin{abstract}
    Coverage motion planning is essential to a wide range of robotic tasks. Unlike conventional motion planning problems, which reason over temporal sequences of states, coverage motion planning requires reasoning over the spatial distribution of entire trajectories, making standard motion planning methods limited in computational efficiency and less amenable to modern parallelization frameworks. In this work, we formulate the coverage motion planning problem as a statistical inference problem from the perspective of flow matching, a generative modeling technique that has gained significant attention in recent years. The proposed formulation unifies commonly used statistical discrepancy measures, such as Kullback-Leibler divergence and Sinkhorn divergence, with a standard linear quadratic regulator problem. More importantly, it decouples the generation of trajectory gradients for coverage from the synthesis of control under nonlinear system dynamics, enabling significant acceleration through parallelization on modern computational architectures, particularly Graphics Processing Units (GPUs). This paper focuses on the advantages of this formulation in terms of scalability through parallelization, highlighting its computational benefits compared to conventional methods based on waypoint tracking.
\end{abstract}

\IEEEpeerreviewmaketitle

\section{Introduction}

Coverage motion planning---the problem of synthesizing a robot trajectory to visit regions of the space based on certain specifications (e.g., density maps)---is essential to a wide range of robotic tasks, including autonomous exploration~\cite{kim_plgrim_2021,cao_representation_2023}, manipulation~\cite{shetty_ergodic_2022,bilaloglu_tactile_2025}, and embodied learning~\cite{prabhakar_mechanical_2022,berrueta_maximum_2024}. For example, a UAV performing a search and rescue mission must plan a trajectory that comprehensively searches across regions of the space based on prior information, such as satellite images~\cite{murphy_human-robot_2004}. Compared to conventional motion planning problems, coverage motion planning faces two unique computational challenges: (1) it requires reasoning over longer horizons; (2) it involves reasoning not only over the temporal specification of the task (e.g., reaching a sequence of states) but also over the spatial specification (e.g., reaching a set of spatial landmarks). The first challenge suggests that coverage motion planning could benefit from modern computational architectures, in particular, parallelization. However, the need to reason over both temporal and spatial specifications makes existing methods---such as those generating a reference trajectory by solving a traveling salesman problem (TSP)~\cite{xie_integrated_2019, peng_visual_2020, shah_multidrone_2020, cao_representation_2023}---difficult to parallelize.

\begin{figure*}
    \centering
    \includegraphics[width=0.99\textwidth]{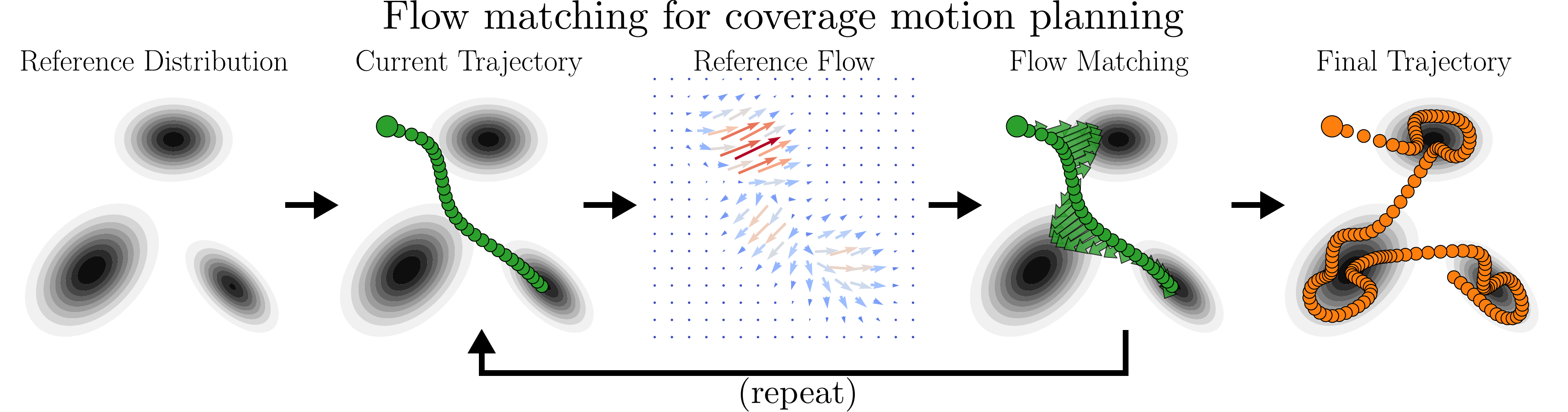}
    \caption{Our method adapts flow-based generative inference methods to generate dynamically feasible flow directions on the state trajectory. Reference flow on the state trajectory is generated using standard methods from machine learning and accelerated through GPU parallelization. Lastly, we synthesize control gradients that generate dynamically feasible flow on the state trajectory by solving a linear quadratic regulator (LQR) problem.}
    \label{fig:overview}
    \vspace{-1em}
\end{figure*}

In the era of parallelism, a new formulation of the coverage motion planning problem is necessary to better integrate parallelization techniques. Therefore, we propose viewing coverage motion planning as a statistical inference problem, where the goal is to generate a trajectory---as a set of samples constrained by the robot's dynamics---with the same statistical properties as a reference spatial distribution specifying the coverage task (see Fig.~\ref{fig:overview}). The proposed formulation~\cite{sun_flow_2025}, which stems from a branch of robotic motion planning techniques named ergodic control~\cite{miller_ergodic_2016}, separates the coverage motion planning problem into two steps: first, a gradient vector field is generated based on the statistical discrepancy between the trajectory and the reference distribution; then, the robot’s control is synthesized based on the generated gradient, taking into account the robot's dynamics. The first step is equivalent to standard practices in modern generative inference methods, which can be effectively accelerated through parallelization, particularly on Graphics Processing Units (GPUs). The second step is a standard optimal control problem, which can be efficiently solved without the need for parallelization. Essentially, our method focuses on tracking the gradient of the trajectory toward a reference distribution instead of directly optimizing the trajectory given reference waypoints. 

This paper serves as a complementary study for~\cite{sun_flow_2025} focusing on the advantages of the proposed method in terms of parallelization on GPUs. We introduce the formulation and algorithm description, specifically with two approaches for specifying the reference gradient---based on the Stein variational gradient descent~\cite{liu_stein_2016} and based on Sinkhorn divergence~\cite{feydy_interpolating_2019}---highlighting the computational advantage of our method, with and without GPU acceleration, and compared with a baseline method based on the traveling salesman problem (TSP). 

\section{Methodology}

\subsection{Problem formulation}

Denote a system's state as $s(t){\in}\mathcal{X}$, the control as $u(t){\in}\mathcal{U}$, and the continuous-time dynamics as $\dot{s}(t){=}f(s(t){,}u(t))$. The reference distribution (also called target distribution) is denoted as $q(x)$, the domain of which is the same as the robot state space $\mathcal{X}$. We define the \emph{empirical distribution} of the trajectory as:
\begin{align}
    p_{s}(x) = \frac{1}{t_f} \int_0^{t_f} \delta(x-s(t)) dt,
\end{align} where $t_f$ is the trajectory horizon and $\delta(x)$ is a Dirac delta function. 

Given a statistical discrepancy measure $D$, such as the Kullback-Leibler divergence, we can formulate the coverage problem as the following trajectory optimization problem:
\begin{gather}
    \argmin_{u(t)} D(p_s(x) , q(x)), \label{eq:obj} \\
    \text{s.t., } s(t) = s_0 + \int_0^t f(s(\tau), u(\tau)) d\tau.
\end{gather} 

The reference trajectory $q(x)$ specifies the coverage task and its representation can vary depending on the task. For example, $q(x)$ can be presented as a set of samples, which makes it compatible with conventional waypoint-based coverage task specifications. On the other hand, $q(x)$ can also be represented as a continuous utility function specifying the varying importance of different regions across the search space~\cite{luo_adaptive_2018, benevento_multi-robot_2020, chen_adaptive_2024}.

Directly solving the optimization problem (\ref{eq:obj}) is challenging as the statistical discrepancy (\ref{eq:obj}) between the trajectory empirical distribution and the reference distribution is often not compatible with standard trajectory optimization formulas (e.g., time integral of runtime cost functions defined at a specific time). We now introduce the algorithm from ~\cite{sun_flow_2025} that solves the optimization problem in (\ref{eq:obj}) by adapting the flow matching~\cite{lipman_flow_2022} formula from generative modeling.

\subsection{Flow-based statistical inference}

Given a set of $n$ samples $\mathbf{s} {=} \{s_i\}_{n}$ with underlying distribution $p_{\mathbf{s}}(x)$, a vector field $g(x)$, called the \emph{flow vector field}, can be evaluated at each sample as $\delta s_i {=} g(s_i)$. Taking an infinitesimally small step $\epsilon$ along this vector field yields a new set of samples $\{s_i + \epsilon \cdot \delta s_i\}$. In flow-based sample generation, the flow vector field $g(x)$ is constructed iteratively such that the underlying distribution of the new samples converges to the reference distribution $q(x)$ under the discrepancy measure $D(p_{\mathbf{s}}, q)$ (see Fig.~\ref{fig:overview}). Such flow-based sampling generation techniques are widely used in statistical inference and generative modeling, such as in Stein variational gradient descent and optimal transport. We specify two kinds of flow vector fields here.  

\noindent\textbf{[Stein variational gradient flow] } We specify the flow vector field given the set of samples $\{s_i\}_{n}$ as:
\begin{align}
    g(s_i) & = \frac{1}{n} \sum_{j=1}^{n} \left[ k(s_j, s_i) \nabla_{s_j} \log q(s_j) + \nabla_{s_j} k(s_j, s_i) \right], \label{eq:stein}
\end{align} where $k(s,s^\prime)$ is a kernel function that is often specified as a radial basis kernel function in practice. It is shown in~\cite{liu_stein_2016} that this vector field is the steepest descent direction of the KL-divergence between $p_{\mathbf{s}}(x)$ and $q(x)$ in a reproducing kernel Hilbert space. 

\begin{figure}[t!]
    \centering
    \includegraphics[width=0.49\textwidth]{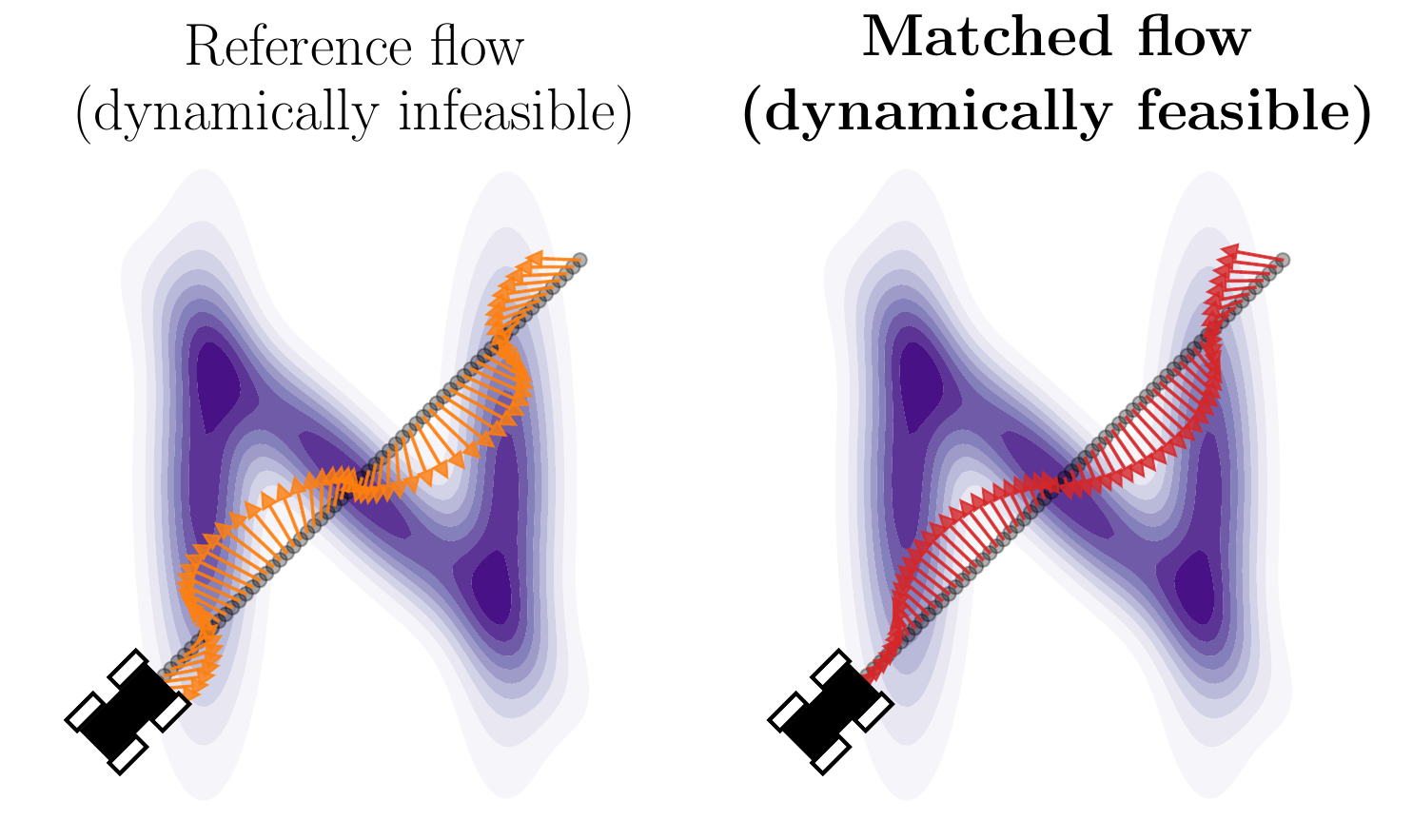}
    \vspace{-2em}
    \caption{Our linear quadratic flow matching formula generates dynamically feasible flow on the state trajectory that closely matches the reference flow.}
    \label{fig:lqr_flow}
    \vspace{-1em}
\end{figure}

\noindent\textbf{[Sinkhorn divergence gradient flow] } We first introduce the entropic regularized optimal transport distance $OT_{\omega}$ between the sample empirical distribution $p_{\mathbf{s}}(x)$ and the reference distribution $q(x)$, where the reference distribution $q(x){=}p_{\mathbf{z}}(x)$ is also represented as a set of $m$ samples $\mathbf{s}^\prime{=}\{s_j^\prime\}_{m}$:
\begin{gather}
    OT_{\omega}(p_{\mathbf{s}}, q) = \min_{T\in\mathbb{R}^{n\times m}} \sum_{i,j} T_{i,j} {\cdot} c(s_i,s_j^\prime) + \omega {\cdot} T_{i,j} \log(T_{i,j}) \nonumber 
    \\
    \text{s.t. } T{\cdot}\mathbf{1}_m = \mathbf{1}_n, T^\top{\cdot}\mathbf{1}_n = \mathbf{1}_m, T_{i,j} \geq 0,  \label{eq:ent_ot}
\end{gather} where $c(s,s^\prime)$ is a cost function that is often specified as the squared norm in practice, and $\omega$ is the entropic regulation weight. Based on the entropic regularized optimal transport distance, we specify the statistical discrepancy measure in our trajectory optimization formula (\ref{eq:obj}) as the Sinkhorn divergence:
\begin{align}
    D(p_{\mathbf{s}}, q) = OT_{\omega}(p_{\mathbf{s}}, q) - \frac{1}{2} OT_{\omega}(p_{\mathbf{s}}, p_{\mathbf{s}}) - \frac{1}{2} OT_{\omega}(q, q). 
\end{align} Importantly, the optimization problem in (\ref{eq:ent_ot}) is convex and can be solved efficiently using the Sinkhorn algorithm~\cite{cuturi_sinkhorn_2013}, the process of which is differentiable with respect to the samples $\mathbf{s}{=}\{s_i\}_n$. Therefore, the gradient flow vector field for Sinkhorn divergence is evaluated as:
\begin{align}
    g(s_i) = \nabla_{s_i} \left[ OT_{\omega}(p_{\mathbf{s}}, q) {-} \frac{1}{2} OT_{\omega}(p_{\mathbf{s}}, p_{\mathbf{s}}) {-} \frac{1}{2} OT_{\omega}(q, q) \right], \label{eq:sinkhorn}
\end{align} which can be efficiently calculated in practice through auto-differentiation.

\begin{figure}[t!]
    \centering
    \includegraphics[width=0.49\textwidth]{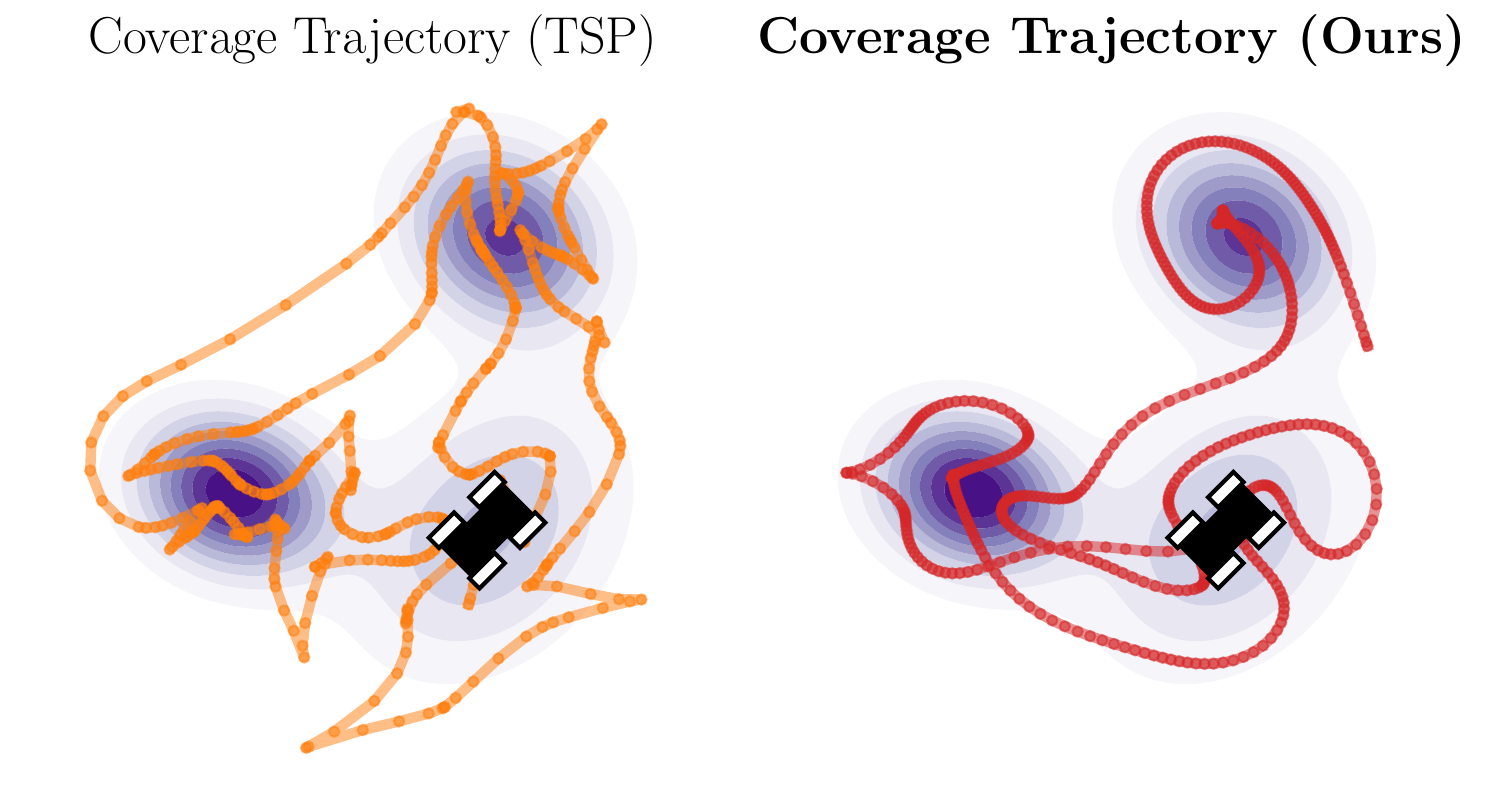}
    \vspace{-2em}
    \caption{Example coverage trajectories generated by the TSP baseline and our flow matching method based on the Stein variational gradient flow for a differential-drive robot.}
    \label{fig:stein_trajs}
    \vspace{-1em}
\end{figure}

\noindent\textbf{[Reference flow on state trajectory]} Given the robot's current trajectory $s(t)$, we can discretize the continuous-time trajectory into discrete time steps $\{t_i\}$, which can be viewed as a set of state samples indexed by the time steps $\{s_{t_i}\}$. Therefore, we can generate a \emph{reference flow} on the state trajectory, denoted as $a(t)$, by evaluating the flow vector field $g(x)$ across the time steps, where $a_{t_i}{=}g(s_{t_i})$. 

\noindent\textbf{[Parallelized flow evaluation]} Crucially, since the robot trajectory is simply viewed as a set of samples, the evaluation of the flow vector field---which essentially only involves summations of various calculations across all the samples---for both the Stein variational gradient flow (\ref{eq:stein}) and the Sinkhorn divergence gradient flow (\ref{eq:sinkhorn}) can be significantly accelerated through parallelization, especially on GPUs and for long-horizon trajectories~\cite{le_accelerating_2023,koide_megaparticles_2024}. Furthermore, since the evaluation of the reference flow does not need to take into account the robot's dynamics, the evaluation can be conducted simultaneously across all the time steps, leading to additional improvements in computational efficiency from parallelization.

\noindent\textbf{[Properties of reference flows]} The Stein variational gradient (\ref{eq:stein}) and the Sinkhorn divergence gradient (\ref{eq:sinkhorn}) have different properties that make them better choices for different coverage tasks. From the coverage performance perspective, the Stein variational gradient only requires access to the score function (the derivative of the log-likelihood) of the reference distribution, thus is more robust when the reference distribution is not normalized~\cite{sun_flow_2025}. Sinkhorn divergence gradient leads to better coverage accuracy on reference distributions with non-smooth and irregular support. We refer the readers to~\cite{sun_flow_2025} for more details on how different gradient evaluation formulas lead to better coverage performance and their computational efficiency.

\begin{figure}[t!]
    \centering
    \includegraphics[width=0.49\textwidth]{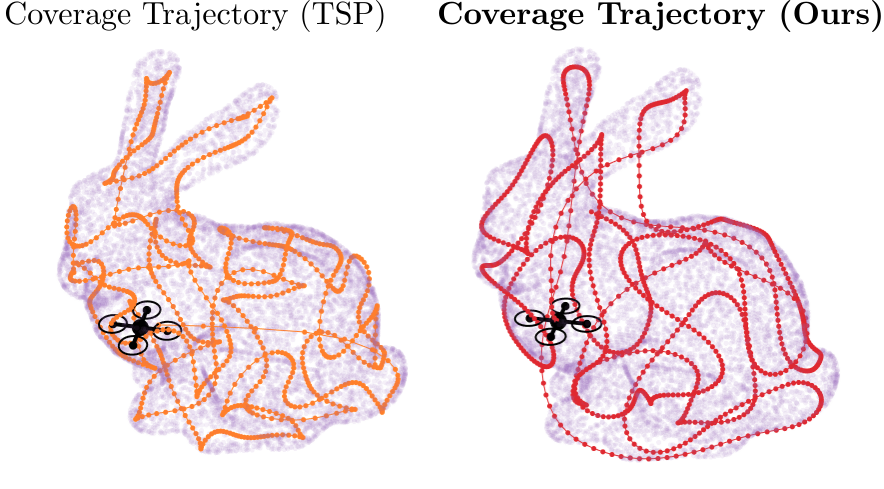}
    \vspace{-2em}
    \caption{Example coverage trajectories generated by the TSP baseline and our flow matching method based on the Sinkhorn divergence gradient flow for an aircraft robot.}
    \vspace{-1em}
    \label{fig:sinkhorn_trajs}
\end{figure}

\subsection{From gradients to control synthesis}

\noindent\textbf{[Intuition] } We cannot directly update the robot trajectory $s(t)$ in the direction of the reference flow $a(t)$, as $s(t)$ is constrained by the dynamics of the system while the evaluation of $a(t)$ is not aware of the dynamical constraints. Instead, we need to synthesize the gradient $v(t)$ on the control $u(t)$ of the system, such that the resulting dynamically feasible flow $z(t)$ on the state trajectory $s(t)$---constrained by the system dynamics $f(s(t), u(t))$---closely matches the reference trajectory gradient $a(t)$. Importantly, there exists a linear dynamics structure between the gradient on the control $v(t)$ and the dynamically feasible flow on the state trajectory $z(t)$~\cite{hauser_projection_2002,miller_trajectory_2013}:
\begin{gather}
    \dot{z}(t) = A(t) z(t) + B(t) v(t), \quad z(0) = 0, \label{eq:linear_dyn} \\
    A(t) = \nabla_{x} f(x(t), u(t)), \quad B(t) = \nabla_{u} f(x(t), u(t)). \nonumber
\end{gather} 

\noindent\textbf{[Linear quadratic flow matching] } Based on the linear dynamics structure in (\ref{eq:linear_dyn}), we can synthesize the gradient on the control by solving the following linear quadratic regulator (LQR) problem:
\begin{gather}
    v(t)^* = \argmin_{v(t)} \int_{0}^{t_f} \vert a(t) - z(t) \vert^2_{Q} + \vert v(t) \vert^2_{R} dt, \label{eq:lqr}\\
    \dot{z}(t) = A(t) z(t) + B(t) v(t), \quad z(0) = 0, 
\end{gather} which can be solved in closed-form using the continuous-time Riccati equation~\cite{hauser_projection_2002} (see Fig.~\ref{fig:lqr_flow}). We can iteratively optimize the robot trajectory to synthesize its statistical property by iteratively solving the LQR problem (\ref{eq:lqr}) and updating the control following the optimal control gradient $v(t)^*$.

\section{Experiment}

\subsection{Benchmark design}

We design the benchmark to evaluate the computational efficiency of our flow matching formula with and without GPU parallelization, as well as a baseline method based on solving the traveling salesman problem (TSP). We implement our formula for both the Stein variational gradient flow and the Sinkhorn divergence gradient flow. 

\begin{figure}[t!]
  \centering
  \includegraphics[width=\linewidth]{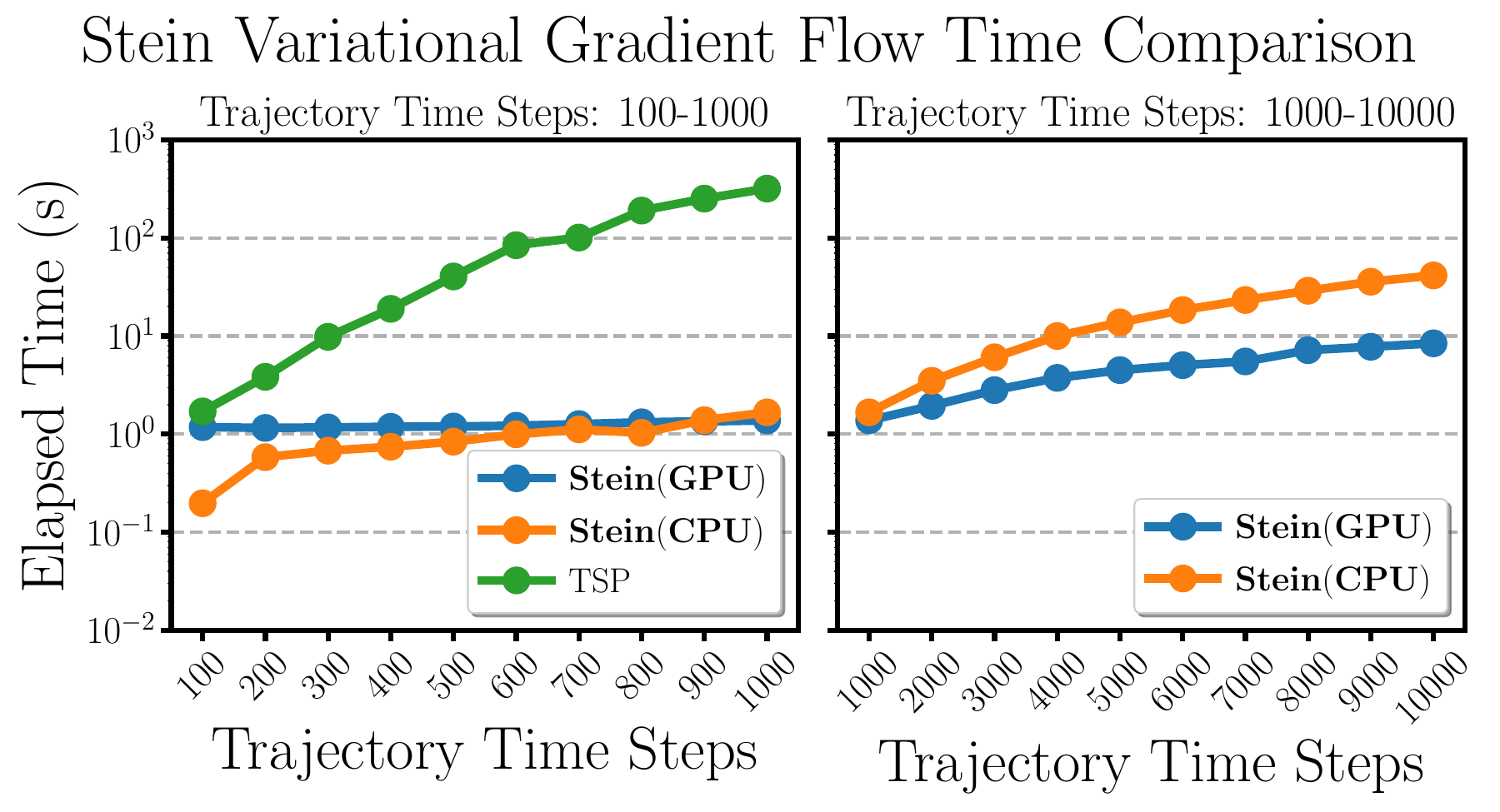}
  \vspace{-1em}
  \caption{Time efficiency comparison of our method using the Stein variational gradient flow (on GPU and CPU) and the TSP baseline. The TSP baseline is not tested beyond 1000 time steps due to the high computation time.}
  \vspace{-1em}
  \label{fig:stein_time}
\end{figure}

To benchmark our implementation with the Stein variational gradient flow, we specify the robot dynamics as a differential drive system and vary the number of time steps of the planning horizon from 100 to 1000 with an interval of 100, and from 1000 to 10000 with an interval of 1000. To benchmark our implementation with the Sinkhorn divergence gradient flow, we specify the robot dynamics as a 3D aircraft. We vary the number of time steps of the planning horizon from 100 to 500 with an interval of 100, and from 500 to 2500 with an interval of 500. We record the elapsed time of each method.  

\subsection{Implementation details}

We implement our method in JAX~\cite{bradbury_jax_2018} for GPU acceleration. We use the OTT package\footnote{\url{https://github.com/ott-jax/ott}} for evaluating the Sinkhorn gradient flow and the LQRax package\footnote{\url{https://github.com/MaxMSun/lqrax}} for solving the LQR problem. All the benchmark experiments are conducted on Intel Xeon w9-3495X CPU and NVIDIA RTX 6000 GPU.

For the TSP baseline, we sample as many points from the reference distribution as the trajectory horizon, order them using heuristic local search via the python-tsp package\footnote{\url{https://github.com/fillipe-gsm/python-tsp}}, and use the result as a reference trajectory for standard trajectory tracking control.

\begin{figure}[t!]
  \centering
  \includegraphics[width=\linewidth]{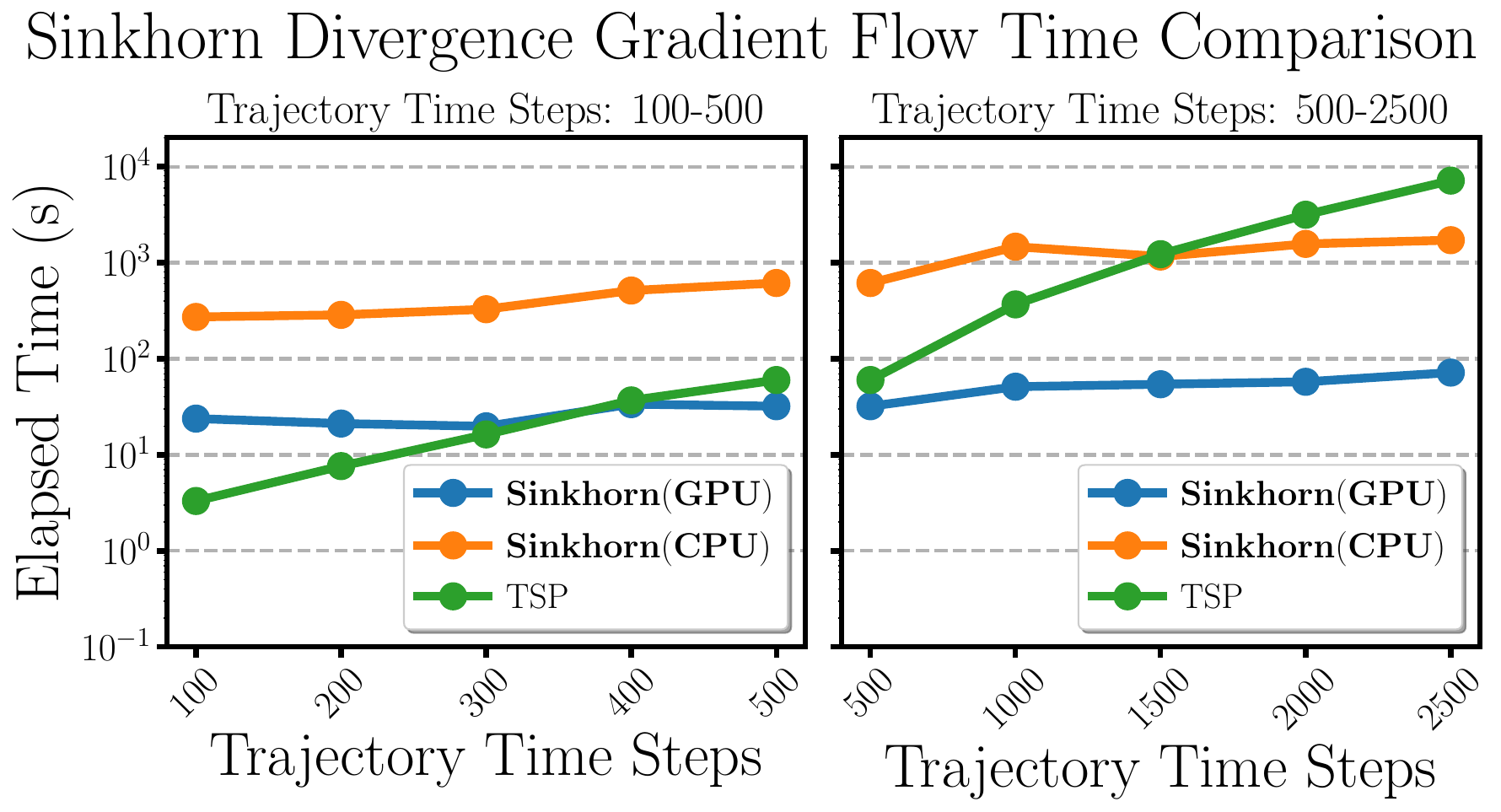}
  \vspace{-1em}
  \caption{Time efficiency comparison of our method using the Sinkhorn divergence gradient flow (on GPU and CPU) and the TSP baseline.}
  \vspace{-1em}
  \label{fig:sinkhorn_time}
\end{figure}

\subsection{Results} Qualitative results of the coverage trajectories from our method (using both the Stein variational gradient flow and the Sinkhorn divergence gradient flow) and the TSP baseline are shown in Fig.~\ref{fig:stein_trajs} and Fig.~\ref{fig:sinkhorn_trajs}. Quantitative results on the computation time across different trajectory horizons can be found in Fig.~\ref{fig:stein_time} for our method using the Stein variational gradient flow and in Fig.~\ref{fig:sinkhorn_time} for our method using the Sinkhorn divergence gradient flow. While our method on CPU and the TSP baseline show lower computation time with shorter time horizons (e.g., less than 300 time steps), our flow matching method with GPU parallelization exhibits better scalability across different time horizons, significantly outperforming other methods at longer time horizons, with both specifications of the reference flow. All methods achieved consistent coverage across the trials, with quantitative coverage accuracy and robustness metrics of our methods available in~\cite{sun_flow_2025}.

We would like to point out that our statistical inference-based formulation of the coverage motion planning naturally leads to multiple equally good optima based on different initial conditions---similar to how different random seeds lead to different but equally good sets of samples. This property can be further leveraged in practice to improve robustness~\cite{lee_stein_2024}.

\section{Conclusion}
In this work, we present a scalable approach to coverage trajectory synthesis by formulating the problem as statistical inference via flow matching. Our method separates the computation of trajectory gradients from control synthesis, enabling significant acceleration through GPU parallelization compared to conventional waypoint-based methods. We demonstrate that our method significantly improves scalability compared to conventional waypoint-based methods, without compromising coverage quality. Furthermore, our method is not mutually exclusive from the waypoint-based methods, which can be used to accelerate the convergence of our method by providing a better initial trajectory. This study highlights the potential of leveraging modern generative modeling techniques and parallel computing for long-horizon robotic planning tasks. Our future work will focus on integrating the proposed method in practical robotics applications, such as large-scale exploration in unstructured environments.

\section*{Acknowledgments}

This work is supported by ARO grant W911NF-19-1-0233 and W911NF-22-1-0286. The views expressed are the authors' and not necessarily those of the funders.

\bibliographystyle{plainnat_titlelink}
\bibliography{references}

\end{document}